\documentclass{amia}
\usepackage{lipsum} %Remove if not needed
\usepackage{multirow}
\usepackage{amssymb}
\usepackage{amsmath}

\begin{document}

\title{Fairly Predicting Graft Failure in Liver Transplant for Organ Assigning}

\author{Sirui Ding$^1$, Ruixiang Tang$^2$, Daochen Zha$^2$, Na Zou, PhD$^1$, Kai Zhang, PhD$^3$, Xiaoqian Jiang, PhD$^3$, Xia Hu, PhD$^2$ }

\institutes{
    $^1$ Texas A\&M University, College station, TX, USA; $^2$Rice University, Houston, TX, USA; $^3$University of Texas Health Science Center, Houston, TX, USA.
}

\maketitle

\section*{Abstract}
%Liver transplant is an essential therapy performed for severe liver diseases. However, donated livers have always been a scarce resource for allocations. This feature makes the assigning of livers a very crucial process. %Prediction or evaluation of some medical metrics serves as the necessary steps during this process. 
%Hospitals usually do the allocative decisions according to a sickest-first policy, which ranks patients using the Model for End-stage Liver Disease (MELD) score. While the MELD score indicates the mortality of the patient, it ignores the features of organs or donors, which can lead to injudicious assigning decisions. In addition, there are few studies that evaluate the MELD score from the fairness perspective that patients under similar medical status should have the same chance of receiving the treatment. Our preliminary experiments show that the MELD sore is prone to predict XXX (unfairness problem). To bridge the gap, we propose a XXX framework that considers features from both organs and donors. Specifically, a XXX network is used to XXX (details of the proposed framework). Besides, we improve the model fairness using a two-steps debiasing method. Experiments are conducted to analyse current unfairness issues and demonstrate the superiority of our method in both prediction and fairness performance. 
%Some method are proposed in traditional clinical way, e.g., MELD score, etc.

\textit{Liver transplant is an essential therapy performed for severe liver diseases. The fact of scarce liver resources makes the organ assigning crucial. Model for End-stage Liver Disease (MELD) score is a widely adopted criterion when making organ distribution decisions. However, it ignores post-transplant outcomes and organ/donor features. These limitations motivate the emergence of machine learning (ML) models. Unfortunately, ML models could be unfair and trigger bias against certain groups of people. To tackle this problem, this work proposes a fair machine learning framework targeting graft failure prediction in liver transplant. Specifically, knowledge distillation is employed to handle dense and sparse features by combining the advantages of tree models and neural networks. A two-step debiasing method is tailored for this framework to enhance fairness. Experiments are conducted to analyze unfairness issues in existing models and demonstrate the superiority of our method in both prediction and fairness performance.}

%  In some circumstances, it is a determinant method to the life of patients.
%  A desirable decision is expected to save as many people as possible and extend their life time to the largest extent. 
% So the clinician experts design some strategies or principals to guide the assigning process of the organs.
\section{Introduction}
Liver transplant is an effective treatment option for end-stage liver diseases and acute liver failure such as hepatic failure. However, the transplant organ resources are scarce compared with the number of patients on the waiting list~\cite{abouna2008organ,saidi2014challenges}. Hence organ assignment becomes a crucial decision that demands careful consideration. A prevalently used assigning strategy is based on the Model for End-stage Liver Disease (MELD) score, which estimates the patient's current status based on three lab test results, including serum creatinine, total bilirubin, and INR of prothrombin time~\cite{wiesner2003meld}. A higher MELD score indicates a worse situation of a patient, and thus a higher priority of the patient to receive organs. The new version MELD score also includes serum sodium for calculation ~\cite{biggins2006meldna}. For pediatric patients, the score definition is different, called Pediatric End-stage Liver Disease (PELD) score~\cite{mcdiarmid2004peld}. We do not differentiate those metrics in our study.

%Some methods are proposed to modify and improve the original MELD score e.g.,~\cite{merion2003longitudinal,myers2013revisionmeld,leise2011revisedmeld,biggins2006meldna}. Through analyzing the MELD score, we find two main drawbacks. 
%Besides pursuing higher living rate or survival length, fairness is also ignored in this medical task. It should be guaranteed the patients, with different demographic background (such as genders, races, etc.) but under similar medical status, should have the same chance of receiving the treatment~\cite{good2005culture,egede2006race,hamberg2008gender}. The two limitations of MELD score may cause unfairness issue in liver transplant.
% To verify this, we conduct preliminary experiments from the statistical perspective. In the left part of Figure~\ref{fig:orrgfr}, we can find there exists obvious gap between each race subgroups' organ receiving rate and graft failure rate. 
%  Moreover, we calculate the Pearson correlation between the MELD score and these two rates in Table~\ref{pearson}.
% The weak correlation with liver transplant statistical metrics and the unfairness issue of MELD score motivate

Despite its prevalence, MELD score has two main drawbacks. First, MELD score does not explicitly consider the post-transplant outcome~\cite{wiesner2003meld,silberhumer2006meldpost}, which is an important metric for organ distributing decisions. Our experimental results show that MELD score only has a very weak correlation with graft failure rate (i.e., the likelihood of graft failure occurs) across genders and races with a Pearson correlation of only $0.36653$ (see Table~\ref{pearson}). Second, MELD score ignores the features of organs and donors~\cite{wiesner2003meld,biggins2006meldna}, which may lead to injudicious organ assigning decisions. (detailed in Section~\ref{sec:52}). As such, researchers are motivated to propose various substitute assignment strategies for liver transplant~\cite{merion2003longitudinal,myers2013revisionmeld}.

% various tasks like computer vision, text processing, etc. Meanwhile, with the development of electronic healthcare record(EHR), many medical tasks can be formulated as problems applicable for machine learning~\cite{caruana2015intelligible,shickel2017deepehr,tjoa2020surveyxaimedical}. So it is with the 
% where some specific machine learning models have been devised for the organ transplant task to partially alleviate these problems,

Machine learning (ML) has provided data-driven solutions for the organ transplant task to better model post-transplant outcomes. The key idea is to train an ML model that takes the features of patients and donors as input, and outputs the predicted outcomes such as pre-transplant mortality, post-transplant mortality, etc. Then, the trained model is deployed to predict a score for each patient-donor pair, which can help clinicians make decisions of organ transplant. Recently, various ML models have been deployed and show promises in the organ transplant task~\cite{delen2010machine,yoon2017personalized,berrevoets2020organite}. For example, Byrd et al.~\cite{byrd2021predicting} use logistic regression and gradient boosting models to predict mortality in liver transplant. Lau et al. apply neural network and random forest to predict graft failure after transplant~\cite{lau2017machine}. Berrevoets et al. propose an interpretable method for real-time organ allocation~\cite{berrevoets2021learningqueueing}. 

% There are also efforts~\cite{delen2010machine,rana2008survival,lau2017machine} applying machine learning model to post-transplant outcome prediction, e.g.,

%Similar to other medical data, sparse features and dense features are the main types of organ transplant data. Though, previous works applying machine learning on organ transplant treat this fact in a trivial way. How to effectively learn from both sparse features and dense features becomes the key point for precise prediction.

% ~\cite{vokinger2021mitigating,norori2021addressing,gianfrancesco2018potential}

% Unfortunately, recent studies suggest that ML models could be unfair and show bias against certain groups of people. For instance, the prediction accuracy on some subgroups, e.g., minority of races, are obvious lower than the general expected accuracy, which can lead to biased organ transplant. However, fairness problem in organ transplant is understudied compared to other tasks, such as prediction accuracy, survival rate, etc. Some preliminary works are discussing this problem~\cite{bertsimas2020balancing,parent2017fair,kaufman2013fairness}, however, most of them haven't approached the unfairness issue from computational perspective or machine learning perspective.

Unfortunately, recent studies suggest that ML models could be unfair and show bias against certain groups of people in organ transplant. Several previous studies have discussed such fairness issues~\cite{bertsimas2020balancing,parent2017fair,kaufman2013fairness}. For example, Byrd et al.~\cite{byrd2021predicting} show that the scores predicted by ML models underrate the mortality of the female group. Our preliminary experiments also show that the gap between GBDT's positive prediction rates across different race groups can be as large as $0.637$ (see Table~\ref{overallresults}). The unfair predictions may cause unfair decisions towards specific race groups. Although some pioneer works point out the unfair issue, there exists no concrete solution that can tackle such unfairness problem to the best of our knowledge. Thus, we are motivated to study the following research question: \emph{can we develop an ML model that is both accurate and fair for the liver transplant task?}

%If we want to widely adopt a model into real-world medical scenario, it has to be both accurate and fair. In practice, improving one of the fairness or accuracy will always sacrifice the other one. Thus, how to balance the accuracy and fairness in organ transplant still remains as a research question. Recently, the fairness topic has gained increasing attention and efforts in ML community. 

%Many computational methods to achieve fairness on machine learning models have been proposed~\cite{du2020fairness,mehrabi2021survey}. However, these methods are infeasible to be directly applied in the organ transplant task. For most of the advanced debiasing methods focus on the fairness issue in neural network~\cite{dong2021individual,xu2021robust}. But most EHR data is tabular data where tree-based models like LightGBM has wide application with promising performance~\cite{jiang2020exploring,chang2019machine,zhang2020predicting}. For these tree-based models, pre-processing and post-processing are more feasible debiasing methods. But unable to end-to-end training makes these methods inconvenient to deploy update online.

While fairness problems in machine learning have been widely investigated recently ~\cite{du2020fairness,mehrabi2021survey}, there are few attempts to study the fairness problem in organ transplant tasks. Developing a fair ML system with competitive accuracy for organ transplant remains a challenging task due to two roadblocks. Firstly, organ transplant datasets contain both dense features (e.g., numerical lab test results) and sparse categorical features (e.g., blood type of recipients and donors). For sparse features, the existing studies simply use one-hot encoding for transformation~\cite{bishara2021machine}. However, one-hot encoding could lead to unsatisfactory performance when the feature cardinality is high due to the curse of dimensionality~\cite{indyk1998approximate}. Secondly, it is challenging to incorporate fairness goals into the training process. Prior work mainly adopts tree-based models~\cite{shaikhina2019decision,sapir2020seeing} for organ transplant prediction due to its strong performance on handling dense inputs. However, existing bias mitigation algorithms mainly focus on the training process \cite{du2020fairness}, including loss design and representation learning ~\cite{wan2021modeling,caton2020fairness}; neither of them can be directly applied to tree-based models because of the indifferentiable property.

% Unfairness issue also exists in machine learning models, which will lead to biased organ transplant as well. Fairness problem in organ transplant is understudied compared to other metrics like prediction accuracy, accumulated rewards of assigning, etc. Some preliminary works are discussing this problem~\cite{bertsimas2020balancing,parent2017fair,kaufman2013fairness}, however, most of them haven't approached the unfairness issue from computational perspective or machine learning perspective. If we want to widely adopt a model into real-world medical scenario, it has to be both accurate and fair. However, how to balance the accuracy and fairness in organ transplant still remains as a research question. Recently, the fairness topic has gained increasing attention and efforts in machine learning community. Many computational methods to achieve fairness on machine learning models have been proposed~\cite{du2020fairness,mehrabi2021survey}. However, these methods are infeasible to be directly applied in the organ transplant task . Currently, most of the advanced debiasing methods focus on the fairness issue in neural network~\cite{dong2021individual,xu2021robust}. But most EHR data is tabular data where tree-based models like LightGBM has wide application with promising performance~\cite{jiang2020exploring,chang2019machine,zhang2020predicting}. For these tree-based models, pre-processing and post-processing are more feasible debiasing methods. But these methods make the framework inconvenient to deploy online.

To tackle these challenges, we propose a fair ML framework for liver transplant. Specifically, we focus on the prediction of liver\footnote{Liver and organ are considered exchangeable in this work when the context has no ambiguity.} transplant graft failure which is one of the most important post-transplant outcomes. Motivated by the strong performance of DeepGBM~\cite{ke2019deepgbm} in recommendation tasks, we use an embedding layer to handle the sparse features and a distillation network with distilled knowledge from a tree-based model to handle the dense features. This design can not only combine the advantages of tree-based models and deep neural networks in handling the sparse and dense features, but also enable us to apply in-processing debiasing techniques to achieve fairness. In particular, we devise a two-step debiasing strategy that mitigates the fairness issues in both the knowledge distillation stage and the end-to-end training stage. We demonstrate the superiority of our framework through extensive experiments on the Standard Transplant Analysis and Research (STAR) dataset. Empirical results show that the proposed framework can precisely and fairly predict graft failure across different races and genders.

% This template should be used as a starting point for AMIA submissions.  A number of Word styles, all beginning with the word ``AMIA", are available for use in your submissions.

% It is important to review the AMIA Call for Participation where types of submissions considered and general requirements for each submission type are listed. All submissions must conform to the format and presentation requirements described in the CFP and at the submission site.

\section{Background of Fairness in Liver Transplant}
In this section, we first describe fairness problems in liver transplant. Then we quantify the unfairness using the fairness metrics adopted in the ML community.

% \textbf{Background.} In the organ transplant assigning scenario, we usually have a waiting list of recipients. The available organs will come into stream and assigned to the patient on waiting list~\cite{berrevoets2021learningqueueing,yoon2017personalized,berrevoets2020organite}. Each time there is an available organ emerging, we need to immediately choose the suitable patient to receive this organ due to the perishable characteristic of organs, like the liver. However we may have multiple patients suitable to receive this organ. In order to make the final but essential decision, we need to rank their priority based on some principles or scores. For example, we can use MELD score, which is a well known clinical scoring method in real-world application. 

\textbf{Fairness of liver transplant.} 
% A fair liver distribution system should give the patients on waiting list under the same medical condition with the same chance of receiving a suitable organ~\cite{good2005culture,egede2006race,hamberg2008gender}. Currently the medical condition is evaluated by calculating the MELD score. Therefore, whether MELD score can fairly represent the status of a patient is very important. However, we find the MELD score has weak correlation with the organ receiving rate of the patient on waiting list from the statistical analysis on our dataset. Directly applying machine learning model will also trigger the unfairness for two reasons. The first is bias may exist in the training data generated from human expertise and records, where the bias comes from human decisions. The second is machine learning model itself will be biased for pursuing higher prediction accuracy while sacrificing the fairness across some subgroups.
Following the existing fairness research in medical fields~\cite{good2005culture,egede2006race,hamberg2008gender}, we study fairness in liver transplant at the group level and focus on race groups and gender groups. Specifically, a fair graft failure predictor should allow patients of different races and genders to have an equal chance of receiving compatible organs. However, fairness is a subjective term so that equal chance could have different interpretations. In this work, we consider fairness defined from two perspectives. On one hand, we expect the patients in different groups to have an equal percentage of being predicted as \emph{graft failed}. In this sense, the patient in different groups will tend to equally receive an organ if allocating organs based on the predicted score. On the other hand, ML models are expected to provide an equal prediction quality for different groups, which can be quantified by true positive rates and false positive rates of the graft failure prediction.

\textbf{Fairness metrics.} The above two fairness definitions correspond to two commonly used fairness metrics for ML models: demographic parity and equalized odds, where the former demands different groups to have an equal percentage of a positive outcome, and the latter requires equal true positive and false positive rates. Specifically, we follow previous work and quantify the degrees of demographic parity and equalized odds with demographic parity difference (DPD) and equalized odds difference (EOD)~\cite{bird2020fairlearn}, respectively. We put detailed mathematical definitions in Section~\ref{sec:3}.

% There are two essential steps, organ assigning decision and post-transplant outcome evaluation. In this work, we approach the fairness issue in both steps. In the organ assigning step, fairness means all the patients with same medical conditions have the same chance of receiving an organ regardless of demographic attributes, e.g., races, genders, etc. In the post-transplant evaluation, we want the post-transplant outcome between subgroups being close to each other. For example if mortality rate or graft failure rate is obvious higher in some subgroups, it may be caused by the unfairness in organ assigning. Because these subgroups may have less chances assigned more suitable organs for them. Though these two fairness definition varies for different steps in organ transplant. They have the same goal that is the organ resource distribution process should be equally suitable and reasonable for all the demographic subgroups under exactly the same assigning rules or principles.

\section{Data and Problem Description}
\label{sec:3}
\textbf{Dataset.} The Standard Transplant Analysis and Research (STAR) organ transplant dataset is collected from patients registered on the Organ Procurement and Transplantation Network (OPTN) waiting list, de-identified by removing all the identifiers from data and randomly shifted dates under IRB protocol approval (HSC-MS-13-0549). It consists of the biomedical information of both patients and organs/donors. The patients include the ones on the waiting list and the recipients who received organ transplants. The dataset also provides follow-up records of recipients' post-transplant outcomes. For the graft failure prediction task, we select $160360$ recipients, where $41.8\%$ of them suffer from graft failure. We manually choose $40$ features of recipients and $40$ features from organs/donors. The race and gender of each recipient are marked as sensitive attributes.
%Now we formally describe the task of fair graft failure prediction from the computational perspective.
%we will also introduce the notations and entail the dataset with the learning goal we focus on.

%\textbf{Task description}
%MELD scoring method take several lab test results of the recipient into calculation. It means the ignorance of features from donors or organs side, though they may be considered implicitly and manually in real-world scenarios. From our analysis on the organ transplant data, the MELD score is weakly related to organ receiving rate and graft failure rate. Besides, we find bias existing in these two metrics, which implies current organ assigning policy may have unfairness issue.

%Motivated by the limitations of MELD score, other assigning method considering organ features, post-transplant outcomes and fairness deserves to be researched and proposed. In this work, we focus on the fair graft failure prediction, which is highly related to the post-transplant outcomes. The graft failure prediction is based on both recipient and organ features. Moreover, the prediction probability of graft failure can also serve as an important score to determine the recipient on waiting list who matches the organ. The fair prediction of graft failure will help for fairer organ assigning decision and post-transplant outcomes.

\textbf{Notations.}
We denote scalars as lowercase alphabets (e.g., $x$), vectors as boldface lowercase alphabets (e.g., $\mathbf{x}$), matrices as boldface uppercase alphabets (e.g., $\mathbf{X}$). We represent the liver transplant dataset as $\mathcal{D}=\{(\mathbf{r}_i, \mathbf{s}_i, \mathbf{o}_i, y_i)\}_{i=1}^N$, where $\mathbf{r}_i \in \mathbb{R}^{M_r}$ denotes the features of the recipient (e.g., various kinds of lab test results, etc), $\mathbf{s}_i \in \mathbb{R}^{M_s}$ denotes the sensitive features of the recipient (e.g., the demographic information), $\mathbf{o}_i \in \mathbb{R}^{M_o}$ denotes the features of the organ, and $y \in \{0, 1\}$ denotes the post-transplant outcome describing whether the graft fails or not; here, $M_r$, $M_s$, and $M_o$ are the corresponding feature dimensions, and $N$ is the total number of data points.

%Let $X$ represents the feature space of recipients/patients, $O$ represents the feature space of organs/donors. Following the supervision way, we formulate a dataset $D$ containing observed recipient-organ pairs with their post-transplant outcome $Y$, which is the graft failed or not in this work. For each sample in the dataset, we use the $(X, O)$ to denote the pair of recipient and organ. Besides the medical features $(X, O)$ of patients and organs, we also include the sensitive attributes $S$, e.g., genders, races, etc of the recipients in the dataset. Thus the dataset can be formulated in $D = \{(X_i, O_i), Y_i, S_i\}, i = 1,...,N$.

\textbf{Objective.}
The goal is to train a model that takes $\mathbf{r}_i$, and $\mathbf{o}_i$ as input, such that it can accurately predict $y_i$ and is also fair w.r.t. the sensitive features $\mathbf{s}_i$ in terms of the fairness metrics. Previous studies have shown that, in most times, improving fairness can harm the model performance~\cite{berk2017convex}. Thus, a desirable model is expected to achieve a good tradeoff and maximize prediction performance and fairness simultaneously.

\textbf{Fairness metric definitions.} We adpot two fairness metrics DPD and EOD~\cite{bird2020fairlearn} in our experiments, defined as follow:
\begin{gather}
     %\text{DPD} = \text{diff}_{s} \{E[\hat{y}(\mathbf{r}, \mathbf{o} \mid s)]\}
     \text{DPD} = \text{diff}_{s} P(\hat{y}=1|s),
     \\
     \text{EOD} = \max[\text{diff}_{s} P(\hat{y}=1 \mid s, y=1), \text{diff}_{s} P(\hat{y}=1 \mid s, y=0)],
\end{gather}

where $\text{diff}_s$ specifies the difference between the largest and the smallest value among the ones across all $s$,  $\hat{y}$ is the model prediction, where $y=1$ represents the positive outcome, e.g., graft failed. Specifically, DPD measures the performance gap between the positive outcomes across all groups, while EOD measures the gap between true positive rates or false positive rates based on the confusion matrix across all groups.

%that can precisely and fairly predict the graft failure of a given recipient-organ pair $(X,O)$. The optimization goal can be represented as $L(Y', Y, S)$, where $L$ is the joint loss function of prediction and fairness metrics, $Y'$ is the model prediction. The joint loss can balance the prediction and fairness performance of the model during the learning process.

\section{Methodology}
\begin{figure}[]
  \centering
    \includegraphics[width=1.0\textwidth]{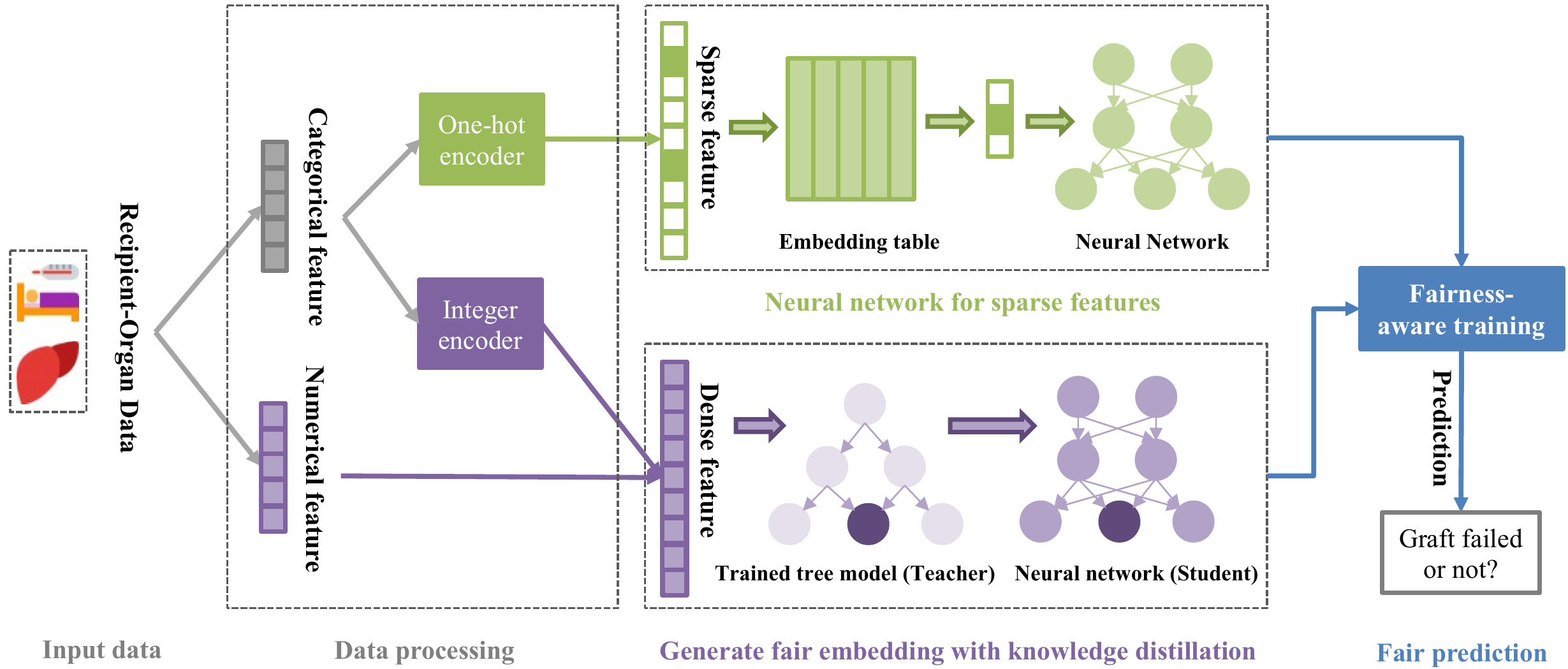}
 \caption{An overview of the workflow for graft failure prediction.}
 \label{fig:workflow}
\end{figure}

In this section, we propose our method for fairly predicting graft failure. Figure~\ref{fig:workflow} shows the workflow, which consists of data processing, prediction model, and fairness-aware training. Firstly, we will introduce how we process the data to extract sparse and dense features from recipients and organs (Section~\ref{sec:41}). Then we introduce a tailored framework that takes the advantage of tree-based models and deep neural networks to make accurate predictions (Section~\ref{sec:42}). Finally, we present a two-step debiasing strategy to achieve fairness (Section~\ref{sec:43}).

\subsection{Data pre-processing}
\label{sec:41}
Following the data pre-processing practice in machine learning, we first impute the missing values. Specifically, we use zeros  to replace the missing values for the numeric data. Then we identify the categorical features (i.e., the features that only have a fixed number of values) and numerical features from the recipient and organ features. For the categorical features, we employ two kinds of encoders, including a one-hot encoder that maps the raw features to one-hot sparse vectors, and an integer encoder which transforms the categorical features into numerical values, where the latter are further concatenated with the original numerical features to serve as the final dense features.

%\textbf{Discussion.} Due to the input contains dense features and sparse features. The challenge for prediction raises here, how can we effectively learn from these hybrid features? From our preliminary experiments, the tree-based model performs good at graft failure prediction task for its strong ability of processing dense numerical features. However, the tree-based model is indifferentiable, which impedes bunches of state-of-the-art in-processing debiasing methods to be applied. Besides, tree-model is less effective faced with sparse features. With the development of deep learning, the neural network(NN) can partly solve these issues. For example, the NN can fit the in-processing debiasing methods and effectively learn from sparse features. But unlike tree-based model, the NN performs unsatisfyingly in our task. Therefore, can we combine these two sorts of method together to make use of their advantages?

\subsection{Combining deep learning and tree-based model for graft failure prediction}
\label{sec:42}
%\textbf{Knowledge Distillation based Prediction Framework}
In previous works, tree-based methods such as random forest~\cite{lau2017machine} have been adopted for graft failure prediction. However, the input space of graft failure prediction consists of both sparse categorical features and dense numerical ones. While tree-based methods often show strong performance on the dense features, they can hardly deal with the sparse features when the feature cardinality is high due to the curse of dimensionality~\cite{indyk1998approximate}. In addition, it is quite difficult to incorporate fairness constraints into the tree-based methods. To tackle these challenges, we propose to combine deep learning and tree-based model for graft failure prediction. Our method is motivated by the success of DeepGBM~\cite{ke2019deepgbm} in recommendation tasks, where an embedding layer and a distillation network with distilled knowledge from a tree-based method are employed to handle the sparse and dense features, respectively. We will first elaborate on how we process the sparse and dense features, and then introduce the end-to-end training objective.

%Fortunately, we are motivated by the task similarity between recommendation system and organ transplant, which are alike in input data and learning goal. Inspired by the design of DeepGBM~\textbf{CITE}, we employ its network design for our prediction task as showed in Figure~\ref{fig:workflow}.

\textbf{Sparse features.} The sparse features from the recipient and the organ are combined and processed by a categorical neural network (CatNN)~\cite{cheng2016wide,guo2017deepfm}, which is  an embedding lookup layer that maps categorical indices to dense vectors, followed by feature interactions. Formally, given a recipient $\mathbf{r}$ and an organ $\mathbf{o}$, we denote the combined sparse features within $\mathbf{r}$ and $\mathbf{o}$ as $\mathbf{x}^{s}$. The embedding of a sparse feature can be denoted as

%We use we have the sparse features $x_s$ and dense features from $x_d$ from concatenated patient and organ features $(X, O)$. We have the embedding in CatNN as follows, where $E_s$ is the output embedding vector of embedding layer with trainable weights $\theta_1$.

\begin{equation}
    E_{\mathbf{V}_j}(x_j^{s}) = \text{embedding\_layer}(x_j^{s}, \mathbf{V}_j),
\end{equation}
where $x_j^{s}$ is the value of the $j^{\text{th}}$ sparse feature of $\mathbf{x}^{s}$, $\mathbf{V}_j \in \mathbb{R}^{c \times d}$ stores all the trainable embedding vectors of the $j^{\text{th}}$ sparse feature, and $c$ and $d$ are the cardinality and the dimension of the embedding table, respectively. Then a factorization machine (FM) is adopted to learn the first/second-order interactions of these features, denoted as $E_{\text{fm}}(\mathbf{x}^s)$, and a deep neural network is applied to learn the higher order interactions of these features, denoted as $E_{\text{deep}}(\mathbf{x}^s)$. For more details of FM and the deep neural network, please refer to Eq.~(2) and Eq.~(3) in~\cite{ke2019deepgbm}. The output of FM and the neural network are summed to obtain the final sparse representations:
%, will take the embedding as their inputs. FM will learn the linear interactions between sparse features and neural network will learn implicit interactions between them, formally represented as

%\begin{equation}
%    E_{fm}(\mathbf{x}^s) = FM(E_s), E_{deep} = DEEP(E_s)
%\end{equation}

\begin{equation}
    y_{\text{CatNN}}(\mathbf{x}^s) = E_{\text{fm}}(\mathbf{x}^s) + E_{\text{deep}}(\mathbf{x}^s)
\end{equation}

\textbf{Dense features.} Similarly, we combine the dense features of recipient $\mathbf{r}$ and organ $\mathbf{o}$, denoted as $\mathbf{x}^d$. To take the advantage of the tree-based models in handling dense features, we train a neural network to distill the knowledge from a trained tree-based model~\cite{ke2017lightgbm}. This is not an easy task because the structures of the trees and neural networks are naturally different. Fortunately, Ke et al.~\cite{ke2019deepgbm} proposes an effective tree distillation strategy by distilling the clustering patterns of the leaf nodes. First, since tree-based methods often do not use all the features but instead greedily choose the useful features, we only select the used features of a tree to train the neural network. Formally, let $\text{NN}_{\text{dense}}$ be the neural network for processing the dense features, $\mathbb{I}$ be the indices of the features that are used in the tree, and $\mathbf{x}^d[\mathbb{I}]$ denote the used dense features. Then $\text{NN}_{\text{dense}}$  will take as input $\mathbf{x}^d[\mathbb{I}]$. Second, we train $\text{NN}_{\text{dense}}$  by distilling the knowledge of how the tree partitions the data. Specifically, a tree-based model essentially partition the data into different clusters, where the data in the same leaf node belong to the same cluster. We train $\text{NN}_{\text{dense}}$ to distill the knowledge from such tree structure by minimizing the following loss function:

\begin{equation}
    L_{\text{KD}} = \sum_{i=1}^{N} \text{mse}(\text{NN}_{\text{dense}}(\mathbf{x}^d_i[\mathbb{I}]), \mathbf{c}_i),
    \label{eq:3}
\end{equation}
where $\mathbf{c}_i$ is the one-hot encoded cluster of the $i^{\text{th}}$ instance, $\text{cross-entropy}(\cdot, \cdot)$ is the cross-entropy loss. Due to the strong expressiveness of deep neural networks, $\text{NN}_{\text{dense}}$ can well approximate the tree structure. Given $\text{NN}_{\text{dense}}$, the dense representations can be obtained by
\begin{equation}
    y_{\text{KD}}(\mathbf{x}^d) = \text{NN}_{\text{dense}}(\mathbf{x}^d_i[\mathbb{I}]) \times \mathbf{q},
\end{equation}
where $\mathbf{q}$ is the leaf values of the tree. For multiple trees, we learn leaf embedding to reduce the dimension of $\mathbf{c}_i$ and group the trees to reduce the number of neural networks following~\cite{ke2019deepgbm}. The leaf embeddings are trained independently based on the tree-based model and will be used as dense representations in the end-to-end training.

%For dense features, the knowledge distillation is leveraged to combine the advantages of tree-based model and neural network. A lightGBM classifier \textbf{CITE} is trained with the data, where the leaf indexes of each tree in classifier are denoted as $\mathbf{L}$.

% \begin{equation}
%     LGBM_{trained} = LightGBMClassifier(x_s, Y)
% \end{equation}

%On account of LightGBM comprises multiple trees, Ke et al.~\cite{ke2019deepgbm} propose a multiple tree distillation method. The first step is using an one-layer MLP to transfer sparse leaf indexes into dense embeddings for efficiency. And this dense embedding will be trained by mapping towards leaf predictions $\mathbf{p_{leaf}}$. The leaf embedding and optimization goal are denoted as:

%\begin{equation}
%    E_{leaf}=MLP_{onelayer}\left( L \right), P_{leaf}=Mapping(E_{leaf})
%\end{equation}

%\begin{equation}
%    \min _{}  \mathcal{L}^{\prime \prime}\left(P_{leaf}, p_{leaf}\right)
%\end{equation}

%Then the knowledge distillation is applied to learn the embedding close to the dense leaf embeddings $E_{leaf}$. Student neural network is employed to achieve this task, where the loss function is usually

%\begin{equation}
%    E_{distillation} = NN(\mathbf{x^s}')
%\end{equation}

%\begin{equation}
%    \mathcal{L}^{\mathbb{T}}=\min \mathcal{L}\left(E_{distillation}, E_{leaf}\right)
%\end{equation}

%The final output of knowledge distilled embedding from tree model can be represented as:

%\begin{equation}
%    y_{KD}(\boldsymbol{x^s}) = Mapping(E_{distillation})
%\end{equation}

\textbf{End-to-end training.} The final output is obtained by combining sparse and dense representations, given as
\begin{equation}
    \hat{y}(\mathbf{x})=\sigma \left(w_{1} \times y_{\text{KD}}(\mathbf{x}^d)+w_{2} \times y_{\text{CatNN}}(\mathbf{x}^s)\right),
\end{equation}
where $w_1$ and $w_2$ are trainable parameters to balance the two representations, $\mathbf{x}$ is combined sparse and dense features from $\mathbf{r}$ and $\mathbf{o}$, and $\sigma(\cdot)$ is the transformation function, such as Sigmoid. Finally, we can train the model in an end-to-end fashion with the following loss:
\begin{equation}
    L=\sum_{i=1}^N \text{cross-entropy}(\hat{y}(\mathbf{x}), y).
    \label{eq:6}
\end{equation}

\subsection{Bias mitigation}
\label{sec:43}
This subsection proposes a two-step debiasing strategy to mitigate the unfairness in the distillation stage and the final training stage, where the former focuses on the bias inherited from the tree-based model when performing knowledge distillation, and the latter aims to achieve fairness in the end-to-end training.

\textbf{Fairness loss.} Motivated by the successes of in-processing methods in debiasing machine learning models~\cite{wan2021modeling,caton2020fairness}, we use fairness loss to incorporate demographic parity in model training. Specifically, we propose the following loss:
%There are standard fairness metrics designed and widely applied like demographic parity, equalized odds, etc. We choose to use demographic parity loss as our fairness constraint.
\begin{equation}
    \text{fairness-loss} (\hat{y}, s_{\text{maj}}) = (E [\hat{y}] -E [\hat{y} | s_{\text{maj}}]) ^2
    %\text{parity-loss} = \left\| \sum_{s_a, s_b \in s} \lvert E \left[\hat{y}(\mathbf{x}) | s_a\right] -E \left[\hat{y}(\mathbf{x}) | s_b\right] \rvert \right\| ^\text{norm}
    \label{eq:7}
\end{equation}
where $\hat{y}$ is the prediction, $s_{\text{maj}}$ is the majority group, $E [\hat{y}]$ is the expected prediction regardless of the sensitive groups, $E [\hat{y} | s_{\text{maj}}]$ is the expected prediction of the majority group. The key idea is to enforce all the sensitive attributes to have similar prediction distributions like the majority group. In training, $E [\hat{y}]$ and $E [\hat{y} | s_{\text{maj}}]$ can be approximated with a batch of data. Thus, Eq.\ref{eq:7} can be naturally applied to the min-batch training of deep learning models.

\textbf{Two-step debiasing.} We propose to debias both the categorical neural network and the network for dense features. In the first step, we achieve fair knowledge distillation by plugging in Eq.~\ref{eq:7} into Eq.~\ref{eq:3}:

\begin{equation}
    L_{\text{KD}} = \sum_{i=1}^{N} \text{mse}(\text{NN}_{\text{dense}}(\mathbf{x}^d_i[\mathbb{I}]), \mathbf{c}_i) + \alpha_{\text{KG}} \times \text{fairness-loss}(y_{\text{KD}}(\mathbf{x}^d), s_{\text{maj}}),
\end{equation}
where $\alpha_{\text{KG}}$ is a hyperparameter to balance prediction performance and fairness. In the second step, we incorporate the fairness constraint into the end-to-end training. Specifically, we similarly debias Eq.~\ref{eq:6} with
%For the second step, we should fairly train the end-to-end model. The training process includes training of the CatNN, so the fairness constraint added to it will also guide the debiasing in CatNN. Similar to generating fair distillation embedding, we also choose demographic parity as our fairness constraint to guide the fair training, the final loss of end-to-end training is denoted as.
\begin{equation}
    L=\sum_{i=1}^N \text{cross-entropy}(\hat{y}(\mathbf{x}), y) + \alpha \times  \text{fairness-loss}(\hat{y}(\mathbf{x}), s_{\text{maj}}),
\end{equation}
where $\alpha$ is a balancing hyperparameter. These two debiasing steps complement each other towards fair final predictions, where the first step focuses on the dense representations which serve as the input of the end-to-end training, and the second step debiases the CatNN and the embedding tables.

\section{Experiment}

\begin{table}[]
\centering
\setlength{\tabcolsep}{10pt}
\begin{tabular}{|c|cc|cc|cc|cc|}
\hline
\multirow{2}{*}{Race} & \multicolumn{2}{c|}{MELD-score}            & \multicolumn{2}{c|}{Number of people}     & \multicolumn{2}{c|}{Receiving rate}        & \multicolumn{2}{c|}{Graft failure rate}    \\ \cline{2-9} 
                      & \multicolumn{1}{c|}{$\,\:$Male$\,\:$} & Female & \multicolumn{1}{c|}{$\,\:$Male$\,\:$} & Female & \multicolumn{1}{c|}{$\,\:$Male$\,\:$} & Female & \multicolumn{1}{c|}{$\,\:$Male$\,\:$} & Female \\ \hline
I                     & \multicolumn{1}{c|}{20.05852}  & 20.36856  & \multicolumn{1}{c|}{89700}     & 49815     & \multicolumn{1}{c|}{0.56405}   & 0.49941   & \multicolumn{1}{c|}{0.32300}   & 0.29592   \\ \hline
II                    & \multicolumn{1}{c|}{21.56156}  & 22.74271  & \multicolumn{1}{c|}{10209}     & 8131      & \multicolumn{1}{c|}{0.60251}   & 0.57004   & \multicolumn{1}{c|}{0.36482}   & 0.34067   \\ \hline
III                   & \multicolumn{1}{c|}{21.14621}  & 21.49130  & \multicolumn{1}{c|}{18282}     & 12074     & \multicolumn{1}{c|}{0.51400}   & 0.47176   & \multicolumn{1}{c|}{0.27754}   & 0.26194   \\ \hline
IV                    & \multicolumn{1}{c|}{17.82069}  & 19.35089  & \multicolumn{1}{c|}{5878}      & 3095      & \multicolumn{1}{c|}{0.53215}   & 0.53312   & \multicolumn{1}{c|}{0.24616}   & 0.25818   \\ \hline
V                     & \multicolumn{1}{c|}{22.13557}  & 23.38609  & \multicolumn{1}{c|}{686}       & 676       & \multicolumn{1}{c|}{0.54082}   & 0.44822   & \multicolumn{1}{c|}{0.28032}   & 0.28713   \\ \hline
VI                    & \multicolumn{1}{c|}{22.20161}  & 19.42105  & \multicolumn{1}{c|}{248}       & 152       & \multicolumn{1}{c|}{0.50000}   & 0.55263   & \multicolumn{1}{c|}{0.26613}   & 0.29762   \\ \hline
VII                    & \multicolumn{1}{c|}{19.80120}  & 20.59470  & \multicolumn{1}{c|}{664}       &  491  & \multicolumn{1}{c|}{0.63253}   & 0.60285  & \multicolumn{1}{c|}{0.26905}   & 0.27703   \\ \hline
\end{tabular}
\caption{Statistical information from liver transplant dataset}
\label{statanalysistable}
\end{table}

In this section, we perform analysis on the datasets and conduct experiments to evaluate the proposed framework. We mainly focus on the following research questions: 
%will first introduce the implementation details of our framework and the experimental settings. Then, we will present and analyse the unfairness issue we discovered in liver transplant. To demonstrate the effectiveness of our method, we compare the method with baseline models on both prediction performance and fairness.
\begin{itemize}
    \item \textbf{RQ1:} Does MELD score align with the post-transplant outcomes for different races and genders (Section~\ref{sec:52})? 
    \item \textbf{RQ2:} Can the proposed framework makes accurate and fair predictions of the graft failure (Section~\ref{sec:53})?
    \item \textbf{RQ3:} How does each stage of debiasing contribute to the fair predictions (Section~\ref{sec:54})?
\end{itemize}

\subsection{Experimental setting} 

%The experiments are conducted on STAR (Standard Transplant Analysis and Research) organ transplant dataset. The dataset consists biomedical features and demographic features of both recipients and donors. From the dataset, we can get the pre-transplant and post-transplant status of the patient on the waiting list. In this work, we focus on the graft failure prediction task, which is a binary classification task highly related to post-transplant mortality.

\textbf{Baselines.} To better demonstrate the effectiveness in prediction performance and debiasing, we choose two categories of baselines. The first is traditional clinical method, which is MELD score in our experiments. We use the Logistic Regression~\cite{fan2008liblinear} as a classifier with MELD score as the input. The second category is machine learning models, where we use Logistic Regression, Random Forest~\cite{breiman2001randomforest} and GBDT~\cite{friedman2001gbdt}.

\textbf{Evaluation protocol.} For a fair comparison, all the machine learning methods are tested under the same setting as follows: the dataset are randomly splited for 5-fold cross-validation for training and testing. The input of all models are the same $80$ features selected based on patient and organ/donor's pre-transplant status. The area under the receiver operating characteristic curve (ROC AUC) is employed as the metric to measure the prediction performance. Meanwhile, two fairness metrics, demographic parity difference (DPD) and equalized odds difference (EOD) are used to evaluate the fairness of prediction towards different groups. 

\textbf{Implementation details.} We implement baseline methods with scikit-learn~\cite{scikit-learn} and the proposed method with PyTorch. The model is trained on Tesla V100 GPU. The knowledge distillation and end-to-end training stages are both trained for 10 epochs. We use Adam~\cite{kingma2014adam} as optimizer for knowledge distillation and AdamW~\cite{loshchilov2017adamw} for end-to-end training. The learning rate is set as $0.001$ in both steps.

\subsection{Statistical analysis of the liver transplant dataset}
\label{sec:52}
For statistical analysis, we select the patients from $7$ main races and $2$ genders with recorded MELD scores. There are $14$ subgroups intersected by races and genders. The average MELD score and the total number of people of each divided subgroup are calculated in Table~\ref{statanalysistable}. We can observe there are obvious gaps between each subgroups' MELD score. The minimum MELD score is only $76.2\%$ of the maximum MELD score. Additionally, the size of majority races is much larger than minority races.

% From the Figure~\ref{fig:sizescore} and Table~\ref{statanalysistable}, we can observe there exists notable gaps of MELD score between different subgroups. The difference between the size of them is more obvious.

Due to the variety existing in each subgroup's MELD score and group size, we take two perspectives that correspond to the organ receiving rate and graft failure rate to better investigate the liver transplant task.
\begin{itemize}
    \item \textbf{Organ receiving rate (ORR)} represents the chance of a group of patients on the waiting list to receive organs. We use the accumulated samples on the waiting list recorded receiving liver transplants based on the MELD score as the number of receiving patients for each subgroup, denoted as $n_r$. The receiving rate is calculated by dividing them by the total number of people in this group registered on the waiting list, denoted as $n_w$. 
    \item \textbf{Graft failure rate(GFR)} reflects the percentage of graft failed for a group of patients who have received the transplant liver. We count the recorded graft failure samples, denoted as $n_f$, and divide it by the number of patients who already received organs, denoted as $n_r$. These two metrics provide an intuitive measure to explore organ assigning and post-transplant outcomes, which are the most essential stages in the liver transplant task. 
\end{itemize}
Formally, these two metrics can be denoted as:

\begin{equation}
    ORR = \frac{n_r}{n_w};
    GFR = \frac{n_f}{n_r}.
    \label{eq10}
\end{equation}

\begin{figure}[]
  \centering
    \includegraphics[scale=0.5]{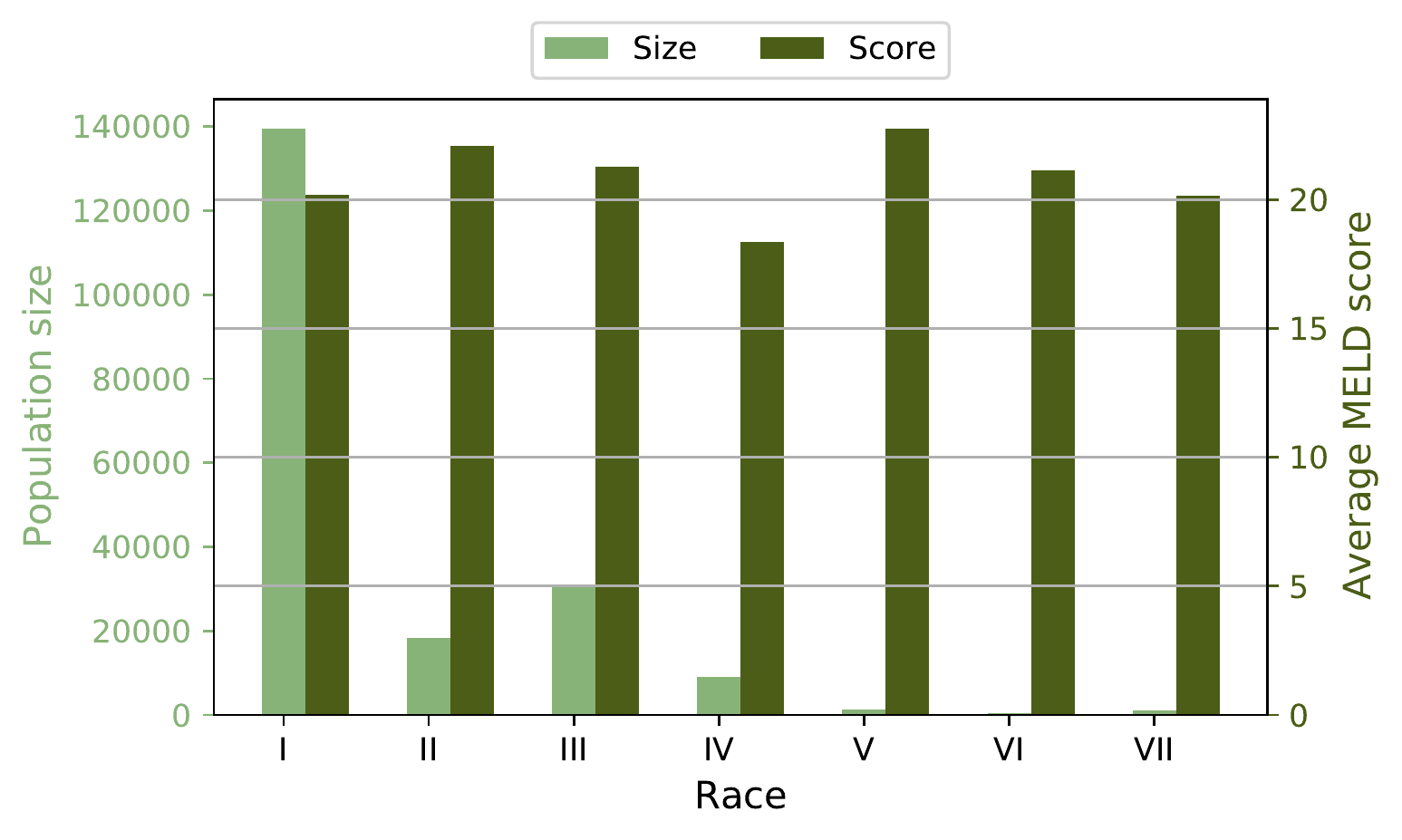}
    \includegraphics[scale=0.5]{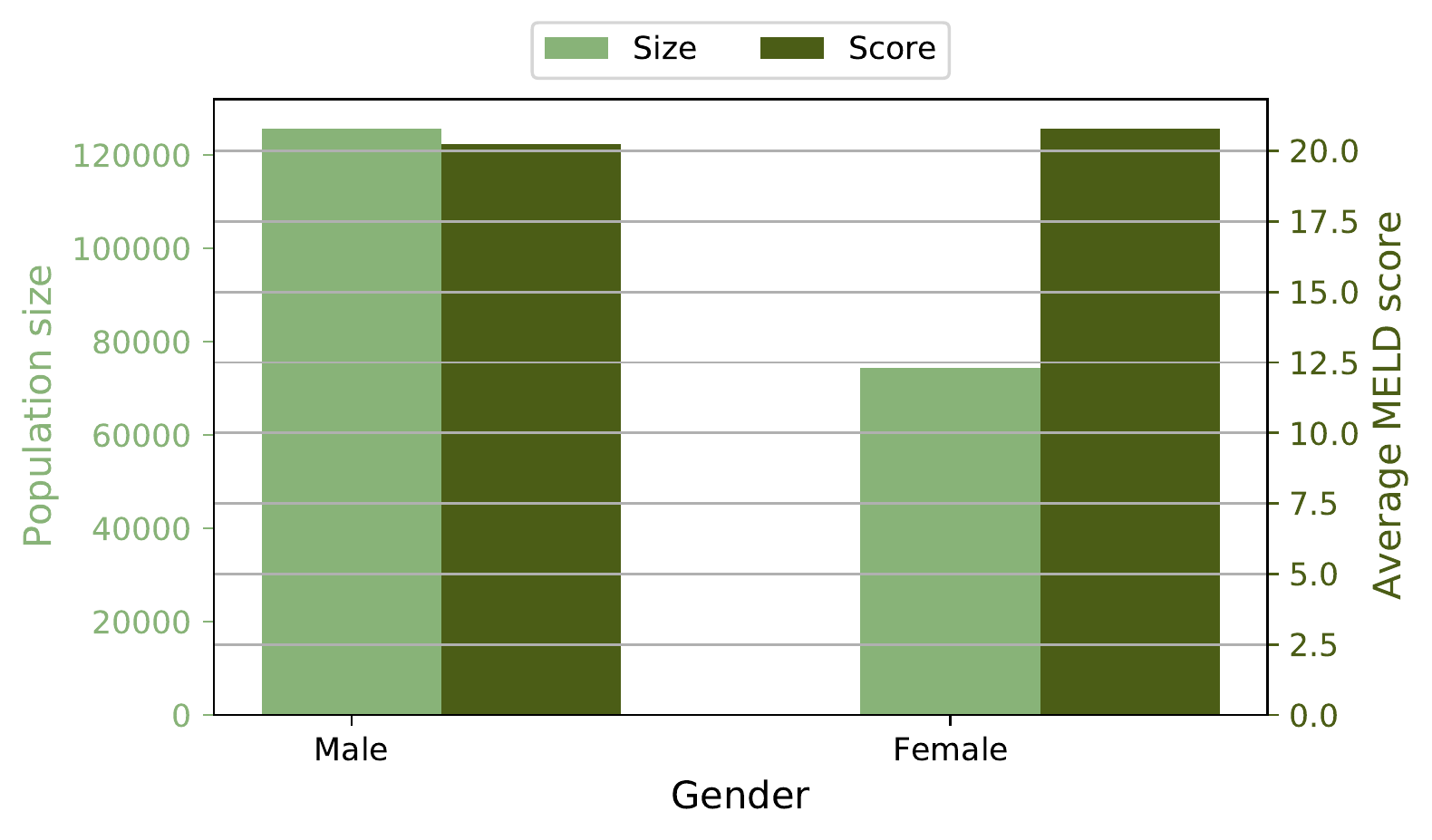}
 \caption{Population size and average MELD score across races and genders.}
 \label{fig:sizescore}
\end{figure}

\begin{figure}[]
  \centering
    \includegraphics[scale=0.5]{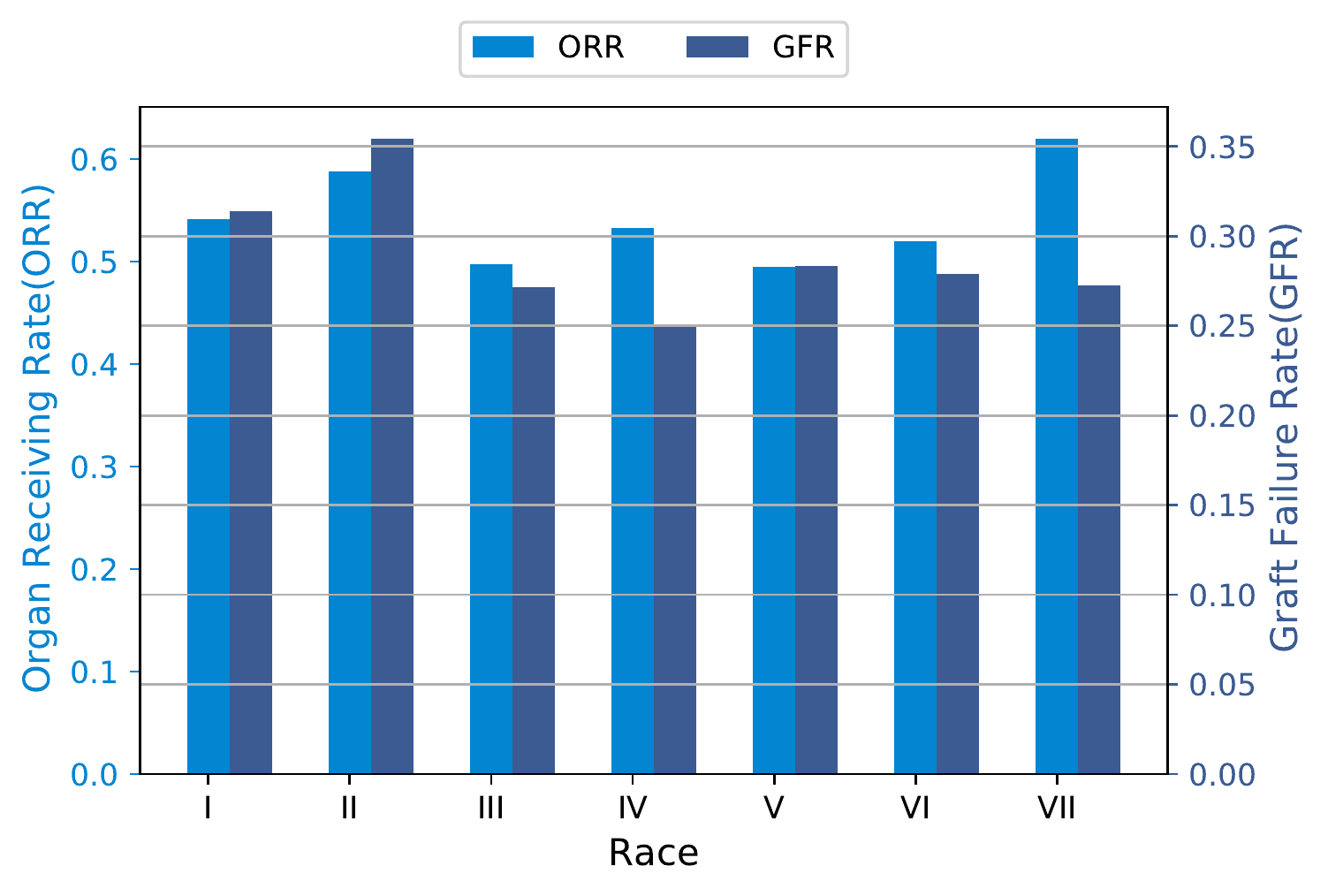}
    \includegraphics[scale=0.5]{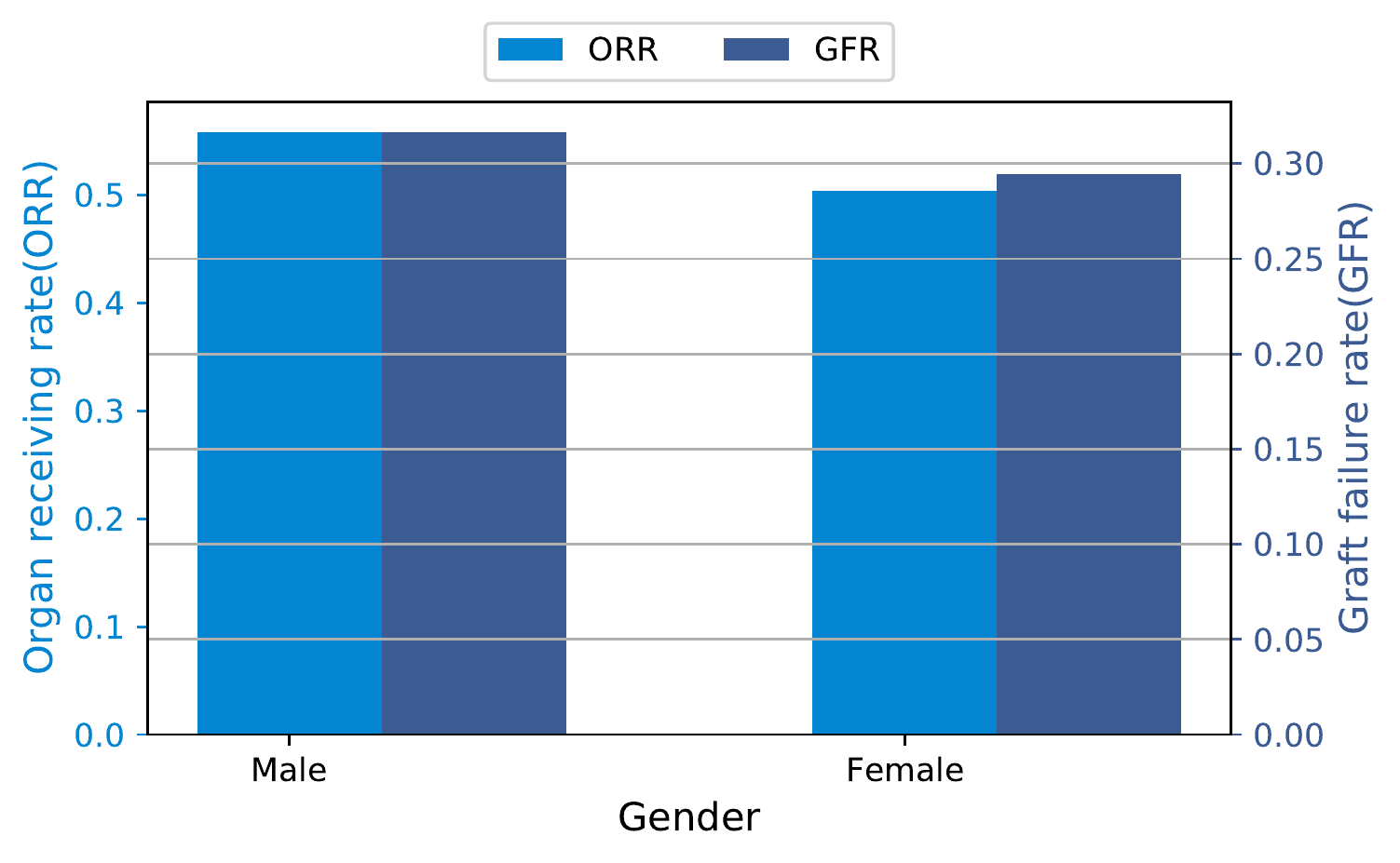}
 \caption{Average organ receiving rate and graft failure rate across races and genders.}
 \label{fig:orrgfr}
\end{figure}

\begin{table}[]
\centering
\begin{tabular}{|c|c|c|}
\hline
                     & MELD-score & Population size \\ \hline
Organ receiving rate & -0.32376   & -0.02243         \\ \hline
Graft failure rate   & 0.36653    & 0.33444         \\ \hline
\end{tabular}
\caption{Pearson correlation between demographic information and liver transplant metrics}
\label{pearson}
\end{table}

For the organ receiving rate, we can observe from the organ receiving rate column in Table~\ref{statanalysistable} that obvious gaps exist between organ receiving rate of different subgroups. The highest receiving rate is $0.63253$ of subgroup interacted by race VII and male, while the lowest receiving rate is $0.44822$ of subgroup interacted by race V and female. However, the latter subgroup's average MELD score is significantly higher than the former subgroup. This means the latter subgroup should have higher priority on the waiting list, which contradicts our findings from the observed data. This phenomenon indicates the MELD score does not align with organ receiving rate. As presented in Table~\ref{pearson}, the Pearson correlation between organ receiving rate and MELD score is $-0.32376$. This means the MELD score has no close relation with the organ receiving rate from the group-level analysis. 

%To briefly summarize, we find unfairness issue laying in organ assigning and it may not be directly caused by MELD-score and size of the subgroups.

For the graft failure rate, we can observe that notable gaps exist between graft failure rates across different subgroups as shown in the graft failure rate column in Table~\ref{statanalysistable}. The subgroup with the highest graft failure rate is the male race II subgroup with a $0.36482$ graft failure rate. The lowest graft failure rate exists in race IV male groups, which is $0.24616$. The MELD score of the latter subgroup is smaller than the former subgroup. It suggests better pre-transplant medical condition, which may explain the lower graft failure rate. To quantify and further look into the relations between MELD score and graft failure rate, we calculate the Pearson correlation between them. The Pearson correlation is still very weak as shown in Table~\ref{pearson}. It implies the MELD-score cannot indicate group-level graft failure rate at the post-transplant stage. 

% In short, unfairness issue exists in graft failure rate and MELD score cannot reflect the graft failure rate.

To summarize, we analyze two main components of organ transplant statistically, the organ assignment and post-transplant outcome. The results show remarkable gaps across subgroups in both two components, which indicates a strong bias existing in organ transplant systems.

% \usepackage{multirow}
% \usepackage{multirow}

\subsection{Results of prediction and fairness performance}
\label{sec:53}
We conduct experiments to compare the prediction and fairness performance of the proposed method with multiple baseline methods (Table~\ref{overallresults}). The key observation is that the proposed model can provide competitive prediction performance with less bias across subgroups.

Compared with the MELD score, we observe that machine learning models show much stronger prediction capability of graft failure. The poor graft failure prediction performance of MELD score aligns with the weak correlations between MELD score and graft failure rate from statistic analysis in Table~\ref{pearson}. The tree model has better and less biased prediction performance than linear model. This may be caused by the tree model's internal selection of features, which could implicitly omit some features with bias.

Compared with baseline machine learning methods, when the sensitive attribute is race, the proposed method can significantly debias the prediction with only $2.1\%$ decreases of ROC AUC, while the two fairness metrics decrease by $5.5\%$ averagely. As for gender, the ROC AUC decreases only $2.0\%$, however, the two fairness metrics decrease by $32.2\%$ on average. Recall that the parity loss we applied is based on the demographic parity. In Table~\ref{overallresults}, we observe improvement not only on DPD but also on EOD. This can validate the effectiveness of our debiasing method, which can generally mitigate the unfairness issues.

\begin{table}[]
\begin{tabular}{|c|ccc|ccc|}
\hline
\multirow{2}{*}{Model}      & \multicolumn{3}{c|}{Sensitive attribute: Race}                            & \multicolumn{3}{c|}{Sensitive attribute: Gender}                    \\ \cline{2-7} 
                            & \multicolumn{1}{c|}{ROC AUC} & \multicolumn{1}{c|}{DPD}       & EOD       & \multicolumn{1}{c|}{ROC AUC} & \multicolumn{1}{c|}{DPD} & EOD       \\ \hline
MELD-score                  & \multicolumn{1}{c|}{0.505$\pm$0.000} & \multicolumn{1}{c|}{---}        & ---        & \multicolumn{1}{c|}{0.505$\pm$0.000}  & \multicolumn{1}{c|}{---}  & ---        \\ \hline
Logistic Regression         & \multicolumn{1}{c|}{0.777$\pm$0.000}        & \multicolumn{1}{c|}{0.648$\pm$0.017}          &    {0.834}$\pm${0.007}     & \multicolumn{1}{c|}{0.777$\pm$0.000}        & \multicolumn{1}{c|}{0.021$\pm$0.000}    &  0.033$\pm$0.001    \\ \hline
Random forest               & \multicolumn{1}{c|}{0.804$\pm$0.000}        & \multicolumn{1}{c|}{0.630$\pm$0.030}          &   0.703$\pm$0.047   & \multicolumn{1}{c|}{0.804$\pm$0.000}        & \multicolumn{1}{c|}{0.020$\pm$0.001}    & 0.036$\pm$0.001        \\ \hline
GBDT                        & \multicolumn{1}{c|}{0.809$\pm$0.000}        & \multicolumn{1}{c|}{0.637$\pm$0.027}      &  0.713$\pm$0.033    & \multicolumn{1}{c|}{0.809$\pm$0.000}        & \multicolumn{1}{c|}{0.017$\pm$0.000}   & 0.031$\pm$0.001  \\ \hline

% W/o debiasing          & \multicolumn{1}{c|}{0.792$\pm$0.001}        & \multicolumn{1}{c|}{\textbf{0.584}$\pm$\textbf{0.033}}          &  0.718$\pm$0.055         & \multicolumn{1}{c|}{0.792$\pm$0.001}        & \multicolumn{1}{c|}{0.012$\pm$0.001}    &   0.025$\pm$0.001        \\ \hline

W/o first-step          & \multicolumn{1}{c|}{0.793$\pm$0.000}        & \multicolumn{1}{c|}{\textbf{0.596}$\pm$\textbf{0.022}}          &  0.687$\pm$0.038         & \multicolumn{1}{c|}{0.792$\pm$0.000}        & \multicolumn{1}{c|}{0.016$\pm$0.002}    &   0.027$\pm$0.002        \\ \hline
W/o second-step & \multicolumn{1}{c|}{0.793$\pm$0.001}        & \multicolumn{1}{c|}{0.616$\pm$0.041}          &   0.745$\pm$0.076      & \multicolumn{1}{c|}{0.793$\pm$0.001}        & \multicolumn{1}{c|}{0.014$\pm$0.007}    &  0.026$\pm$0.009       \\ \hline
Ours          & \multicolumn{1}{c|}{0.792$\pm$0.000}        & \multicolumn{1}{c|}{\textbf{0.597}$\pm$\textbf{0.015}} & \textbf{0.662}$\pm$\textbf{0.029} & \multicolumn{1}{c|}{0.793$\pm$0.001}        &  \multicolumn{1}{c|}{\textbf{0.011}$\pm$\textbf{0.001}}    & \textbf{0.022}$\pm$\textbf{0.003} \\ \hline
\end{tabular}
\caption{Comparison of prediction and fairness performance on graft failure prediction}
\label{overallresults}
\end{table}

\subsection{Ablation study}
\label{sec:54}

To validate the effectiveness of our two-step debiasing strategy, we conduct ablation study to investigate the contribution of each component. From Table~\ref{overallresults}, we observe that by only adding the debiasing method in knowledge distillation step (first step), the proposed model can only improve the DPD metrics. When only debiasing the end-to-end training step, both fairness metrics improve to some extent. The model achieves the best debiasing performance when the two debiasing steps are combined. This is because the knowledge-distilled embedding and end-to-end training are interleaved, which verifies the necessity of the two-step debiasing strategy.
%As a result, debiasing both parts can achieve the best fairness performance.

\section{Conclusion and future work}
This paper aims at fair graft failure prediction for developing unbiased organ assigning strategy. A two-step knowledge distillation framework is built to encourage fair prediction towards different groups while preserving competitive performance. The fair and competitive prediction performance of the whole framework has been experimentally signified on graft failure prediction dataset. In the future, we will investigate and identify more fairness issues such as intersection fairness problem. Furthermore, we will continue designing debiasing methods for liver transplant tasks, fairness problem discovered from the liver transplant task can also inspire research on other organ transplant systems.

\section{Acknowledgements}
This work was supported by NSF grants IIS-1939716 and IIS-1900990. XJ is CPRIT Scholar in Cancer Research (RR180012), and he was supported in part by Christopher Sarofim Family Professorship, UT Stars award, UTHealth startup, the National Institute of Health (NIH) under award number R01AG066749 and U01TR002062.

\makeatletter
\renewcommand{\@biblabel}[1]{\hfill #1.}
\makeatother

% unstr is used to keep citation order
\bibliographystyle{vancouver}
\bibliography{amia}  

\end{document}